\documentclass{article}
\usepackage{spconf,amsmath,graphicx,hyperref,booktabs,amssymb}
\usepackage{xspace}
\usepackage{pifont}
\usepackage{enumitem}
\usepackage{xcolor}

\newcommand{\ours}{SILK\xspace}

\title{Cluster Validation Indices as Self-Supervised Objectives \\
for Text Representation Learning}

\name{\shortstack{
Kishor Kumar Bhaumik$^{1}$ \quad
Nicolas Roque dos Santos$^{1}$ \quad
Neil Shah$^{2}$ \\
Jia Chen$^{1}$ \quad
Evangelos E. Papalexakis$^{1}$
}}

\address{
$^{1}$University of California, Riverside \quad
$^{2}$Pinterest
}

\begin{document}
\maketitle

\begin{abstract}
Self-supervised fine-tuning refines
the embedding space of a pretrained
language encoder without labels.  However, the commonly used approaches are computationally expensive. Specifically, contrastive learning-based methods need multiview data and in-batch negative examples, while negative-free approaches require auxiliary graphs/networks. An interesting question arises: can self-supervised fine-tuning be done without relying on either additional negatives or graph data? To answer this question,
we introduce \ours
(\underline{Sil}houette-guided \underline{K}-means), which trains on a Cluster Validation Index, an
internal measure of cluster quality without using labels.
\ours clusters 
the
corpus and then regresses a simplified silhouette toward a target value. Each
document is compared only against the $k$ cluster centroids, never against other
documents, so the method needs no augmentation, no negative pairs and one view
per document. On BERT-base, \ours trains $1.46\times$ faster per epoch than the
fastest baseline we evaluate and uses $45.4\%$ less peak GPU memory than the
leanest one. Under frozen-encoder linear probing, \ours stays competitive with the best baselines on three downstream tasks. Our code is available at \url{https://github.com/Kishor-Bhaumik/SILK}.
\end{abstract}



\section{Introduction and Related Work}
\label{sec:intro}

Text embeddings underpin retrieval, semantic search, clustering, and
retrieval-augmented generation, where the evidence supplied to a language model
is selected by the geometry of the embedding
space~\cite{lewis2020retrieval}. Pretrained encoders such as
BERT~\cite{devlin2019bert} carry rich contextual information, but masked
language modeling, in general, does not force text embeddings to align with geometrical information, i.e., the resulting space is anisotropic and separates latent semantic categories poorly~\cite{ethayarajh2019contextual}. Self-supervised fine-tuning
addresses this by reshaping the space with unlabeled text.

Contrastive approaches such as SimCSE~\cite{gao2021simcse} and
DiffCSE~\cite{chuang2022diffcse} build two views of each sentence via dropout
masks or edit-based auxiliary tasks and contrast them against in-batch
negatives, with later work~\cite{wu2022esimcse} refining the augmentations and the sampling.
Introducing negatives adds computational cost:
pairwise comparison is quadratic in the batch and
semantically related texts drawn as negatives are actively pushed apart.
Negative-free methods instead prevent collapse architecturally, through feature
decorrelation, asymmetry, or momentum encoders~\cite{grill2020bootstrap};
SCD~\cite{klein2022scd} combines a self-contrastive term with a decorrelation
penalty over two dropout views. Collapse avoidance then becomes an extra
component to design and tune.

This suggests a question that predates deep representation learning: \textbf{\emph{can a
classical clustering criterion serve directly as the self-supervised objective?}}
Cluster Validation Indices (CVIs) score a partition by intra-cluster
compactness and inter-cluster separation with no ground truth. Among these indices, the silhouette
index~\cite{rousseeuw1987silhouettes} encodes exactly the
preference contrastive learning expresses through pairs; same-cluster points
close, different-cluster points far---but does so without constructing any
pairs. CVIs have been used for five decades to \emph{evaluate}
clusterings. Using a CVI itself as a differentiable training loss is recent~\cite{shiao2023carl}.

In the literature, clustering results have been used to guide training neural networks in self-supervised settings. For instance,
DeepCluster~\cite{caron2018deepcluster} trains a CNN on $k$-means pseudolabels and graph methods use clustering to select positives or to optimize cluster quality directly~\cite{shiao2023carl}. CARL-G~\cite{shiao2023carl} trains a GNN with a simplified-silhouette loss, and uses a transductive GCN over a fixed graph generating all node embeddings in a single forward pass. However, neither method is directly applicable to text corpora as documents are independent sequences without explicit graph structure and the activation memory of a Transformer makes a full-corpus computation graph infeasible for corpora of any size. Therefore, it is worth investigating whether CVI-based learning transfers to text data.
\\ 
\begin{figure*}[t]
    \centering
    \includegraphics[width=1.5\columnwidth]{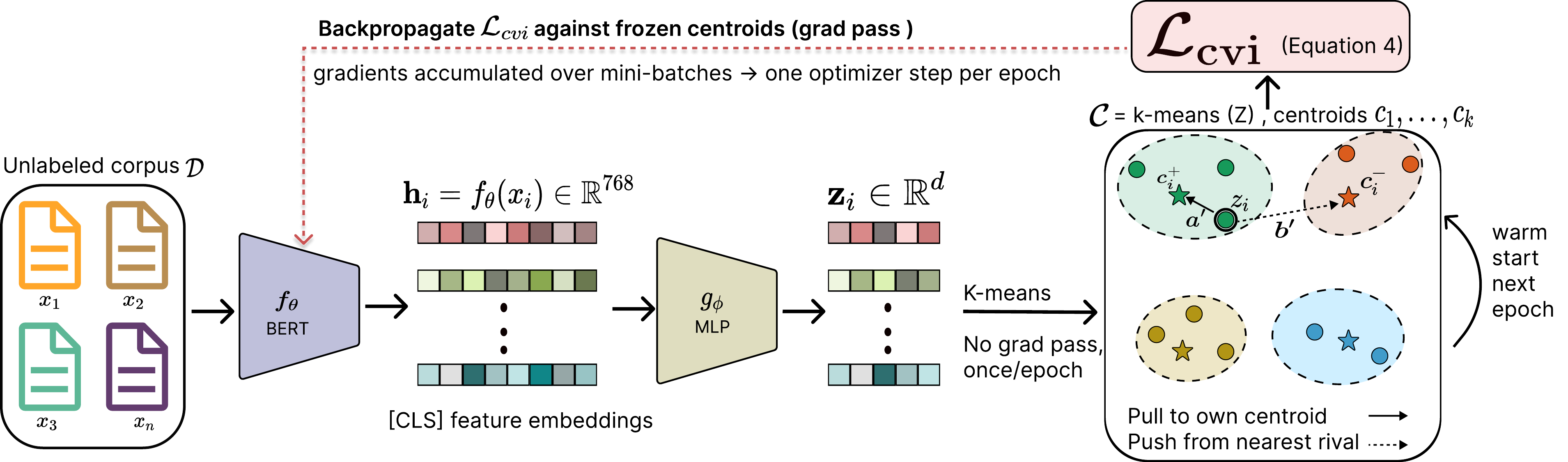}
    \caption{Overview of \ours. Documents $x_1,\ldots,x_n$ are encoded by BERT and projected. $k$-means is fit on the resulting embeddings and the silhouette loss pulls each document toward its assigned centroid while pushing it from the nearest competing one. That CVI loss is backpropagated into BERT. Unlike contrastive methods, no augmented views or negative pairs are constructed.}
    \label{fig:workflow}
\end{figure*}\\ In light of this, we propose \ours (Silhouette-guided K-means). \ours first
embeds an unlabeled corpus of $n$ documents and applies
$k$-means to obtain corpus-level centroids.
It then optimizes a simplified silhouette, which replaces distances to other documents by distances to the $k$ centroids.
The computational cost therefore drops from $\mathcal{O}(n^2)$ pairwise distances to $\mathcal{O}(nk)$ embedding-to-centroid distances, with $k \ll n$. To run this on
a Transformer we separate the scope of the clustering from the scope of the
computation graph, refreshing centroids once per epoch and accumulating
gradients across batches into a single optimizer step. Our contributions are summarized below.

\begin{itemize}[leftmargin=5.0pt, itemsep=2pt]
\item \textbf{CVI loss.}
\ours optimizes cluster quality
directly, contrasting each document against $k$ corpus-level centroids instead
of against other documents. It uses one view per document and requires no augmented views or negative pairs.

\item \textbf{Feasible training scheme.}
A silhouette loss
needs the whole corpus clustered but a Transformer can backpropagate only one
batch at a time. We cluster the corpus once per epoch and accumulate gradients across batches, so peak activation memory scales with the batch rather than with the corpus.

\item \textbf{Lower training cost.} With a single view and no in-batch negatives, per-epoch time and peak GPU memory fall below every baseline we evaluate, while linear-probe accuracy stays competitive with the strongest.
\end{itemize}


\section{Proposed Method}
\label{sec:method}

Figure~\ref{fig:workflow} shows the method. Documents are encoded by BERT, projected, and clustered with $k$-means. The CVI loss on that clustering is backpropagated to fine-tune the encoder.

\subsection{Setup}
\label{ssec:setup_notation}

Given an unlabeled corpus $\mathcal{D}=\{x_1,\dots,x_n\}$ of $n$ documents, let $f_{\theta}$ be
BERT-base with \texttt{[CLS]} pooling and $g_{\phi}:\mathbb{R}^{768}\!\rightarrow\!\mathbb{R}^{d}$
a 2-layer MLP projector with $d\ll 768$. Let $\mathbf{h}_i=f_{\theta}(x_i)$ be the encoder output. The embedding of document $i$ is

\begin{equation}
\mathbf{z}_i=
\frac{g_{\phi}(f_{\theta}(x_i))}{\|g_{\phi}(f_{\theta}(x_i))\|_2},
\qquad i=1,\dots,n ,
\label{eq:proj}
\end{equation}
Normalization keeps a few high-variance directions from dominating the distances, which matters because BERT features are anisotropic~\cite{ethayarajh2019contextual}.
The projector is used only during fine-tuning. At evaluation we discard $g_{\phi}$ and use the $768$-dimensional encoder features~\cite{grill2020bootstrap,shiao2023carl}. Let
$\mathcal{C}=\{\mathcal{C}_1,\dots,\mathcal{C}_k\}$ be a $k$-means partition of
$\{\mathbf{z}_i\}_{i=1}^{n}$, $\mathbf{c}_j$ the centroid of $\mathcal{C}_j$,
$\mathbf{c}_i^{+}$ the centroid of the cluster containing $\mathbf{z}_i$ and
$\mathbf{c}_i^{-}$ the nearest other centroid.

\subsection{Silhouette-Guided Objective}
\label{ssec:sil}

The silhouette index~\cite{rousseeuw1987silhouettes} asks, for each point,
whether it lies closer to its own cluster than to the nearest competing one,
comparing the mean distance to fellow members (compactness) against the mean
distance to the nearest other cluster (separation). This is the geometric
preference contrastive learning encodes through pairs, obtained without pairs
or labels.
The computational cost grows quadratically with the number of documents, $\mathcal{O}(n^2)$, because all pairwise distances
must be calculated
and retained for backpropagation, which is untenable for a Transformer on tens
of thousands of documents.\\We therefore use the \emph{simplified
silhouette}~\cite{hruschka2004simplifiedsilhouette,shiao2023carl}, which
replaces each set of points by the centroid summarizing it:

\begin{equation}
a'(\mathbf{z}_i) = \|\mathbf{z}_i - \mathbf{c}_i^{+}\|_2, \qquad
b'(\mathbf{z}_i) = \|\mathbf{z}_i - \mathbf{c}_i^{-}\|_2,
\label{eq:ab_simple}
\end{equation}\\where $a'$ is how far a document has drifted from its own centroid and $b'$ how
close it has come to its nearest rival. The per-document score is

\begin{equation}
s'(\mathbf{z}_i)=
\frac{b'(\mathbf{z}_i)-a'(\mathbf{z}_i)}
     {\max\{a'(\mathbf{z}_i),b'(\mathbf{z}_i)\}}\in[-1,1],
\label{eq:simsil}
\end{equation}\\near $1$ when the document sits far nearer its own centroid than any other and
negative when it would be better placed elsewhere. Each document is compared
against $k$ centroids rather than $n-1$ neighbours, giving $\mathcal{O}(nkd)$
time and $\mathcal{O}(nk)$ memory.\\What does optimizing Eq.~\eqref{eq:simsil} do to the encoder? In the common case
$a'<b'$ we have $s'=1-a'/b'$, whose partials are $\partial s'/\partial a' =
-1/b'$ and $\partial s'/\partial b' = a'/b'^{2}$. Raising the score therefore pulls the document toward its own centroid and
pushes it from the nearest competing one. The push is weighted by $a'/b'^{2}$, so a
document already tight on its centroid is repelled weakly, while one caught
between two clusters is repelled strongly. The objective contrasts each document against a small set of
corpus-level \emph{summaries} rather than against individual documents, which is
precisely why no negative sampling is required.\\We do not maximize the mean score. A silhouette of $1$ is reached only when
every cluster collapses to a point; since $k$-means inevitably misassigns some
documents, maximization drives those onto the wrong centroid and destroys the
within-cluster variation a downstream classifier needs~\cite{shiao2023carl}. We
regress toward a bounded target $\tau\in[-1,1]$ where the outer sum runs over the $N_B$ mini-batches of an epoch:

\begin{equation}
\mathcal{L}_{\text{cvi}} = \frac{1}{N_B}\sum_{\mathcal{B}} \left| \tau - \frac{1}{|\mathcal{B}|}\sum_{i\in\mathcal{B}} s'(\mathbf{z}_i) \right|,
\label{eq:loss}
\end{equation}

\vspace{-0.5cm}
\subsection{Clustering under a Bounded Computation Graph}
\label{ssec:loop}

CARL-G evaluates its objective in one forward pass per
epoch~\cite{shiao2023carl}, which a transductive GCN permits because all $n$
node embeddings live in a single computation graph. However, Transformer activations
scale as $\mathcal{O}(nLTd)$ in documents, layers, tokens, and width, placing a
full-corpus graph orders of magnitude beyond device memory. Stepping per
mini-batch is no remedy either: $k$-means would then be fit to a few hundred
randomly drawn documents, leaving the centroids of Eq.~\eqref{eq:ab_simple}
sample-dependent and the loss batch-local.\\We decouple the two scopes. An inference-mode pass builds no graph and produces
the embeddings for GPU $k$-means~\cite{johnson2019billion}, initialized with
$k$-means$++$~\cite{arthur2007k} in the first epoch and warm-started from
the previous centroids thereafter, so the loss target does not jump when the
representation has barely moved. A second pass recomputes embeddings with
gradients enabled, evaluates Eq.~\eqref{eq:loss} against the frozen centroids,
and frees each batch's graph before building the next; the accumulated
gradients give one optimizer step per epoch. Every batch is scored against the same frozen centroids, so this step is computed under one consistent clustering,
while peak activation memory falls from $\mathcal{O}(n)$ to
$\mathcal{O}(|\mathcal{B}|)$. We pay one extra forward pass and the graph shrinks
by a factor of $n/|\mathcal{B}|$ where $|\mathcal{B}|$ is the batch size. This is what lets a silhouette objective run on
a Transformer.

\section{Experiments}
\label{sec:exp}

\subsection{Setup}
\label{ssec:setup}

\noindent\textbf{Protocol.}
All methods fine-tune \texttt{bert-base-uncased} on the unlabeled 20
Newsgroups~\cite{lang1995newsweeder} \emph{all} subset ($\sim$18{,}846
documents), with headers, footers and quotes removed. We fine-tune on documents
rather than sentences so the corpus carries the topical structure a clustering
objective can exploit. Afterwards we discard the projector, freeze the encoder and fit a linear
classifier on its \texttt{[CLS]} features. No supervised gradient reaches the
encoder, so the probe measures the geometry of the learned space rather than the
capacity of downstream fine-tuning.

\noindent\textbf{Baselines and tasks.}
We compare against raw BERT (no fine-tuning), SimCSE~\cite{gao2021simcse} and
DiffCSE~\cite{chuang2022diffcse} (contrastive), and SCD~\cite{klein2022scd}
(negative-free, non-contrastive).
We probe on TREC-6~\cite{li2002learning} (6 classes),
Snippets~\cite{phan2008learning} (8 classes), and
StackOverflow~\cite{xu2015short} (20 classes). These span
question types, web snippets, and technical short text, and none matches the
fine-tuning domain, so the probe also measures transfer. All methods, including the baselines, are fine-tuned for 20 epochs.

\noindent\textbf{Hyperparameters.}
\ours uses target $\tau=0.5$, a 2-layer MLP projector with output $d=128$, and a maximum sequence length of $128$ tokens. The batch size is $128$. We train with AdamW
at learning rate $10^{-5}$ and weight decay $10^{-4}$ for $20$ epochs under a
cosine schedule with $10$ warmup epochs. GPU $k$-means runs for at most $100$ assignment and update steps and stops earlier if it converges. The first epoch tries $10$ initializations; later epochs warm-start from the previous centroids and use a single trial. We report the mean and standard deviation over $5$ seeds, each covering both fine-tuning and linear probing. The number of clusters is chosen per downstream dataset by validation-split linear-probe accuracy ($k=150$, $120$, and $200$; see Sec.~\ref{ssec:k}).
The probe trains a single linear layer with AdamW (lr $10^{-2}$, weight decay
$10^{-4}$, batch $512$) for $15$ epochs, and we report final-epoch test accuracy.

\begin{table}[tb]
\centering
\small
\setlength{\tabcolsep}{4pt}
\begin{tabular}{lcccc}
\toprule
 & SimCSE & DiffCSE & SCD & \ours \\
\midrule
Avoids negative pairs        & \ding{55} & \ding{55} & \checkmark & \checkmark \\
Augmentation-free            & \ding{55} & \ding{55} & \ding{55}  & \checkmark \\
Single view per document     & \ding{55} & \ding{55} & \ding{55}  & \checkmark \\
No extra anti-collapse term & \checkmark & \ding{55} & \ding{55} & \checkmark \\
\bottomrule
\end{tabular}
\caption{What each objective requires. \ours satisfies all four properties by
contrasting against centroids instead of documents.}
\label{tab:properties}
\end{table}

\begin{table}[tb]
\centering
\small
\setlength{\tabcolsep}{3pt}
\begin{tabular}{lccc}
\toprule
Method            & TREC-6 & Snippets & StackOverflow \\
\midrule
BERT              & $79.04\pm0.40$ & $71.13\pm0.50$ & $72.76\pm0.30$ \\
SimCSE            & \underline{$90.64\pm0.30$} & \underline{$80.70\pm0.40$} & {\boldmath$79.16\pm0.50$} \\
DiffCSE           & $88.15\pm0.60$ & $80.67\pm0.30$ & $76.18\pm0.40$ \\
SCD               & $89.15\pm0.50$ & $79.35\pm0.40$ & $77.28\pm0.60$ \\
\midrule
\ours (ours)      & {\boldmath$92.47\pm0.40$} & {\boldmath$82.04\pm0.30$} & \underline{$78.24\pm0.50$} \\
\bottomrule
\end{tabular}
\caption{Linear-probe test accuracy (\%, mean $\pm$ std over 5 seeds, each
covering both fine-tuning and probing) on frozen features. All methods
fine-tuned on 20 Newsgroups.}
\label{tab:main}
\end{table}

\subsection{Downstream Accuracy}
\label{ssec:main}

Table~\ref{tab:main} reports linear-probe accuracy. \ours improves over the
strongest baseline by $1.83$ points on TREC-6 and $1.34$ on Snippets, and
trails SimCSE by $0.92$ on StackOverflow. Every fine-tuning method improves
substantially on raw BERT, confirming that the pretrained space is poorly suited
to linear separation and that the gains come from reshaping it rather than from
the probe. The StackOverflow result is the informative one. It has $20$ classes of
technical short text, the furthest of the three from the Usenet discussion
corpus \ours was fine-tuned on, and it is also where a clustering objective has
least to work with: with few tokens per document, the centroids defining $a'$
and $b'$ are estimated from sparse, overlapping topics. A loss built on cluster structure should give ground where that structure is
weak, and it does. This also marks the boundary of where the method applies.

\vspace{-0.5 cm}
\subsection{Training Cost}
\label{ssec:eff}

Table~\ref{tab:efficiency} reports per-epoch time and peak memory under matched
conditions. \ours trains $1.46\times$ faster per epoch than SCD, the fastest
baseline, and uses $45.4\%$ less memory than SimCSE, the leanest---despite
spending an extra forward pass on clustering every epoch. The saving is structural rather than incidental. Contrastive baselines hold two views
of every document in the graph and materialize a
$|\mathcal{B}|\times|\mathcal{B}|$ similarity matrix, so their negatives live
\emph{inside} the batch. \ours keeps one view and contrasts against $k$
centroids that are external to the batch and fixed for the epoch. This is also
why the two respond differently to batching: shrinking the batch alters the
InfoNCE~\cite{oord2018representation} objective by removing negatives, whereas for \ours it changes only how
finely the same corpus-level gradient is accumulated.

\begin{table}[tb]
\centering
\small
\begin{tabular}{lcc}
\toprule
Method & Time/epoch (min) & Peak GPU mem (GB) \\
\midrule
SimCSE  & 3.017 & 21.823 \\
DiffCSE & 4.200 & 33.766 \\
SCD     & 2.578 & 22.536 \\
\ours   & \textbf{1.766} & \textbf{11.917} \\
\bottomrule
\end{tabular}
\caption{Self-supervised training cost per epoch on an RTX A6000, batch size
128, FP32. Time includes the corpus embedding pass and $k$-means as well as the gradient update. Memory is the mean per-epoch peak of the PyTorch allocator; linear probing is excluded. All methods run for the same number of epochs (20), so per-epoch cost is proportional to total cost.}
\label{tab:efficiency}
\end{table}

\vspace{-0.5 cm}

\subsection{Number of Clusters}
\label{ssec:k}

\begin{figure}[tb]
    \centering
    \includegraphics[width=1.0\columnwidth]{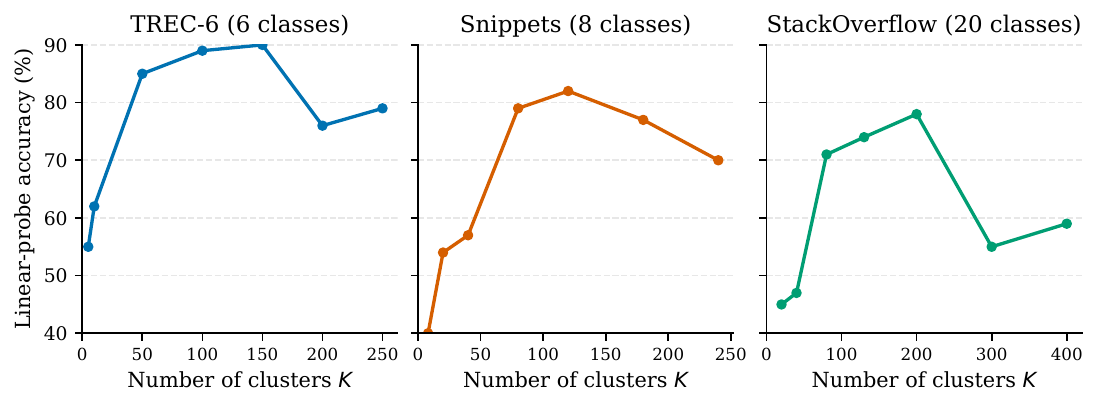}
    \caption{Validation-split linear-probe accuracy against the number of clusters $k$. Each point is a separate training run. The $k$ used in Table~\ref{tab:main} is the peak of this same validation-split search. The best
    $k$ exceeds the downstream class count by an order of magnitude on all three
    tasks.}
    \label{fig:k-sensitivity}
\end{figure}

A natural guess is to set $k$ to the number of downstream classes. We instead treat $k$ as a hyperparameter and, for each downstream dataset, train separate models across a range of $k$. Figure~\ref{fig:k-sensitivity} plots validation-split linear-probe accuracy for that search, and Table~\ref{tab:main} uses the $k$ at each peak. Accuracy climbs
steeply well past the class count, and on all three tasks the best $k$ is
roughly an order of magnitude larger. Beyond that point the gains stop and the
curves turn back down. Both ends of the range are easy to read. Too few clusters force unrelated topics
onto a shared centroid, so $a'$ is minimized by pulling together documents that
should stay apart. Too many split each topic into near-duplicates, leaving $b'$
set by a neighbour that differs from the document's own cluster in name only,
and the separation term stops carrying signal. The same preference for large $k$
appears in clustering-based self-supervision for vision and
graphs~\cite{caron2018deepcluster,shiao2023carl}, which suggests the clusters
work as a partition of the representation space rather than a stand-in for the
label set. In practice the region around each peak is broad, so $k$ only needs
to be set to the right order of magnitude.
\vspace{-0.6 cm}
\section{Limitations and Future Work}
\label{sec:limitations}

We used a small corpus ($\sim$18.8k documents) for clustering in our experiments. On a much larger corpus mini-batch or streaming $k$-means~\cite{sculley2010web} would
take its place. The centroids would then be approximate and whether the gains survive that approximation is the first thing to test. The method also depends on $k$, which must be chosen per corpus. Finally, the objective may be applied beyond text modality. The same recipe should carry over to images and other
modalities. We leave that for future work.
\vspace{-0.5cm}
\section{Conclusion}
\label{sec:conclusion}

We presented \ours in this paper, which fine-tunes a language encoder using a Cluster
Validation Index as its training loss. Each document is compared against cluster centroids alone and a single optimizer step per epoch is enough. To run it on a
Transformer we clustered at the epoch level and accumulated gradients across
batches, which keeps memory bounded while every update uses one consistent clustering. \ours
trains faster and uses less memory than every baseline we evaluate and stays
competitive on linear-probe accuracy. Cluster validation indices have long been
used to judge representations; these results suggest they can also be used to
learn them. 


\noindent\small{\textbf{Acknowledgements.}} The work was partially supported by
the Agriculture and Food Research Initiative Competitive Grant no.
2020-69012-31914 from the USDA National Institute of Food and Agriculture, and
by the National Science Foundation under grant no. 2431569, grant 2524228, and
the CREST Center for Multidisciplinary Research Excellence in CyberPhysical
Infrastructure Systems (MECIS) grant no. 2112650. The views and conclusions in
this paper are those of the authors and should not be interpreted as
representing any funding agencies.

\vspace{4pt}
\noindent\small{\textbf{Compliance with Ethical Standards.}} This work used
publicly available benchmark datasets containing no personally identifiable
information, and required no human or animal subjects. The authors declare no
conflicts of interest.

\bibliographystyle{IEEEbib}
\bibliography{refs}

\end{document}